\documentclass{article}
\usepackage{arxiv} 

\usepackage[T1]{fontenc}
\usepackage[utf8]{inputenc}

\usepackage[sectionbib]{natbib}
\usepackage{bibunits}
\defaultbibliographystyle{dcu}
\defaultbibliography{main}
\setcitestyle{authoryear, open={(}, close={)}}

\usepackage{url}

\usepackage{booktabs}       
\usepackage{amsfonts}       
\usepackage{nicefrac}       
\usepackage{microtype}      
\usepackage{graphicx}
\usepackage{multirow}
\usepackage{import}
\usepackage{MnSymbol}
\usepackage{pifont}
\usepackage{wasysym}
\usepackage{tablefootnote}
\usepackage{array}
\newcolumntype{P}[1]{>{\centering\arraybackslash}p{#1}}

\usepackage{authblk}

\makeatletter

\graphicspath{ {figures/} }

\usepackage{doi}
\usepackage{hyperref}

\title{Advanced Atom-level Representations for Protein \\ Flexibility Prediction Utilizing Graph Neural Networks}

\author[1,2]{Sina Sarparast*}
\author[1]{Aldo Zaimi }
\author[2]{Maximilian Ebert}
\author[2]{Michael-Rock Goldsmith}

\affil[1]{Mila \\ 6666 St-Urbain, 200, Montréal, Quebec, H2S 3H1}
\affil[2]{Computational Molecular Sciences, Congruencetx, 7171 Rue Frederick-Banting, 117, Montréal, Quebec, H4S 1Z9, Canada}

\affil[1]{\textit {\{sina.sarparast,aldo.zaimi\}@mila.quebec}}
\affil[2]{\textit {\{ssarparast,mebert,rgoldsmith\}@congruencetx.com}}

\begin{document}

\maketitle

Protein dynamics play a crucial role in many biological processes and drug interactions. However, measuring, and simulating protein dynamics is challenging and time-consuming. While machine learning holds promise in deciphering the determinants of protein dynamics from structural information, most existing methods for protein representation learning operate at the residue level, ignoring the finer details of atomic interactions. In this work, we propose for the first time to use graph neural networks (GNNs) to learn protein representations at the atomic level and predict B-factors from protein 3D structures. The B-factor reflects the atomic displacement of atoms in proteins, and can serve as a surrogate for protein flexibility. We compared different GNN architectures to assess their performance. The Meta-GNN model achieves a correlation coefficient of 0.71 on a large and diverse test set of over 4k proteins (17M atoms) from the Protein Data Bank (PDB), outperforming previous methods by a large margin. Our work demonstrates the potential of representations learned by GNNs for protein flexibility prediction and other related tasks.


\section{Introduction}

The B-factor is an atomic displacement parameter (expressed in square angstroms, \(\mathring{A} ^2\)) that is typically measured experimentally in proteins by X-ray crystallography (\cite{caldararu2019crystallographic}). The growing interest in B-factor measurement has been motivated by a number of early studies suggesting that regions with higher B-factors could correlate with higher mobility (\cite{sun2019utility}). For instance, high mobility regions are characterized by, higher than average flexibility, higher than average hydrophilicity, and higher net charge (\cite{radivojac2004protein}). Therefore, B-factors can be used in protein science for identifying flexible regions that can potentially be targeted for future drug discovery applications (\cite{sun2019utility}). Publicly available X-ray protein structures are deposited in the \textit{Protein Data Bank} (PDB) (\cite{pdb}), which include B-factor measurements along with other properties.

Accurate methods to predict protein B-factors in the absence of experimental structures remain elusive. Due to the lack of experimental protein structures for many relevant targets (full protein, pathogenic mutations, ligand complexes, etc.) structural models, for example generated using AlphaFold (\cite{alphafold}), are missing B-factors. Accurate prediction of B-factors from structure alone could serve as a fast method to evaluate protein dynamics without running an all atomistic protein dynamics simulations.

In this work, we present a deep learning framework for predicting protein B-factors at the atomic level based on 3D protein structure. A set of diverse graph neural network (GNN) architectures were implemented and adapted to this use case. The architecture of Graph Neural Networks (GNNs) demonstrates promising performance in learning representations for graph data, such as recommendation systems, social networks, and various biological data like gene expression, protein-protein interactions, and metabolic pathways. Consequently, we used GNNs to learn representations for the prediction of B-factors by applying graph-based learning on graphs where nodes represent atoms and edges represent atomic bonds. The best model in this work, \textit{Meta-GNN}, reaches a Pearson Correlation Coefficient of $0.71$ on a test set containing over $4k$ proteins ($17M$ atoms). To the best of our knowledge, this is the first work tackling the use case of \textit{atomic} property prediction on proteins using graph neural networks.


Previous studies have pursued two major research approaches, namely protein representation learning at the residue level and small molecule representation learning at the atomic level. The subsequent paragraphs delve into the specifics of each approach and present the relevant findings from previous research. 

\section{Protein representation learning}

With the success of deep learning, significant advances have been made in the last few years in the field of protein representation learning. Graphs are particularly well suited for encoding protein tertiary structures in the form of a network of residues (the amino acids) represented by graph nodes and bonds represented by graph edges. Protein representation learning models based on GNNs have been able to tackle a great number of challenging tasks, including protein structure prediction from residue sequence (\cite{alphafold}), protein function prediction (\cite{pfp1}, \cite{pfp2}), \cite{pfp3}), protein fold classification (\cite{bios2net}), protein model quality assessment (\cite{proteingcn}), protein interface prediction (\cite{protein}), protein-ligand binding affinity prediction (\cite{predicting}, \cite{deepsite}) and protein-protein binding prediction (\cite{pancino2020graph}). A few approaches have successfully combined the protein sequence with the protein 3D structure in order to obtain more expressive multi-level representations (\cite{chen2022structure}, \cite{gearnet}). Although numerous approaches and tasks have been explored in the literature, there is a scarcity of studies focusing on the atomic-level representation for protein-related tasks. There are two primary reasons that account for this: first, the significant computational resources necessary to encode proteins, which are typically comprised of thousands of atoms each; and second, the majority of protein-related tasks, such as predicting protein properties and binding affinity, operate at a high level (i.e. residue level) and do not necessitate encoding at the atomic level as required for tasks involving small molecules.

\section{Small molecule representation learning} A series of approaches mainly inspired by quantum chemistry have been proposed for prediction and classification tasks on small molecules at the atomic level. For example, \cite{mol1} proposes the use of Convolutional Neural Networks (CNNs) to learn molecular fingerprints on graphs for predictive tasks such as solubility, drug efficacy, and organic photovoltaic efficiency. \cite{mpnn} propose the Message Passing Neural Network (MPN) for molecular property prediction tasks on the quantum chemistry QM9 benchmark dataset. Atomic features such as the atom type, the atomic number, the hybridization and the number of hydrogens are encoded in the input node embeddings. Similarly, \cite{schnet} offers a way to model interactions of atoms in molecules, using interaction maps, while respecting quantum-chemical constraints related to energy conservation. Despite success on small molecules, these approaches do not necessarily scale well for large macromolecules such as proteins, where the number of atoms can easily exceed a few thousand and building the interaction maps can become prohibitively expensive. Moreover, in proteins, sequences of residues combine to form secondary structure components (e.g. alpha helices and beta sheets) via hydrogen bonds, which will then fold to form the tertiary 3D structure via non-covalent bonds (e.g. polar hydrophilic hydrogen interactions, ionic interactions, and internal hydrophobic interactions). Capturing these interactions that are only present at the higher protein structure levels is challenging.

\section{Protein B-factor prediction} Most of the existing work on protein B-factor prediction has been done with standard machine learning algorithms. (\cite{Bramer_2018}) generated 2D images of protein features/encodings via a Multiscale Weighted Colored Graph (MWCG). Random forests, gradient-boosted trees and CNNs were then compared for a blind B-factor prediction task. They made use of local features such as the atom type and amino-acid type for every heavy atom, but also of global features such as the number of heavy atoms in the protein and the resolution of the crystallographic data. (\cite{jing2014research}) considered protein features such as disorder, mutation information, secondary structure and physicochemical and biochemical properties. The B-factor prediction was performed with four different machine learning algorithms: linear regression, REP (Reduced Error Pruning) tree, Gaussian process regression and random forest regression. Finally, (\cite{yuan2005prediction}) trained a support vector regressor to predict a B-factor distribution profile from a protein sequence. The majority of the past work presents two main limitations: (i) their performance rarely exceeds a Pearson Correlation Coefficient (CC) value of 0.6 and (ii) the training/testing experiments are performed over a limited number of proteins. Moreover, none of them explores the potential of GNNs for the use case of protein B-factor prediction.

\section{Methods}

\subsection{Data Description}

\paragraph{} All protein structures used in this work were obtained from the \textit{Protein Data Bank (PDB)} (\url{https://www.rcsb.org}) in \textbf{\textit{.pdb}} format. PDB files were parsed using TorchDrug's \citep{torchdrug} data parsing tool, and RDKit (\url{https://www.rdkit.org}).
Typically, each PDB file contains a protein, which is presented as a 3D atom cloud with x, y, z coordinates, and information about chemical bonds between atoms, and other information such as crystallographic resolution (in the case of X-ray and Cryo-EM structures), the author(s) of the file, etc. Hydrogen atoms are typically not resolved by x-ray or Cryo-EM explicitly and their approximate location can often be calculated.
We have a general set of Apo and Holo proteins. The Apo proteins are a set of proteins that do not have their necessary co-factors or ligands bound to it. Conversely, The Holo state consists of proteins that contain ligands.
In parallel, Kinase proteins of different subfamilies and groups were diversely sampled from the PDB, thus making a test dataset of diverse proteins that are representative of known structures of the kinome (labeled as the Kinase dataset in this work). The Kinases are essential for a wide range of cellular functions, including growth, differentiation, metabolism, and apoptosis (programmed cell death), and hence of prime importance for the drug discovery process. They are particularly important in signal transduction pathways, where they transmit signals from the cell surface to the nucleus, ultimately influencing gene expression and cellular responses. Due to their central role in cell regulation, Kinases have been a focus of extensive research, and they are also important targets for drug development, which is the reason we are interested in evaluating performance on them. 

A quantitative comparison between the two sets can be found in Table \ref{tab:table1}. For both datasets, an initial filtering was performed to remove proteins with erroneous B-factor values (e.g. B-factor of 0, and constant B-factor value across all protein atoms, neither are valid proteins for the purpose of this work). A second filtering involved only keeping proteins with an X-ray crystallographic resolution < $2.5\mathring{A}$.

It is generally agreed that deposited proteins with resolutions larger than $2.5\mathring{A}$ are considered unreliable (\cite{carugo2018large}). This is primarily due to the complex interplay of dynamics and experimental errors present in structures with resolutions beyond this threshold, making it challenging to disentangle these factors.

\subsection{Graph Representation}

\begin{figure}
\begin{center}
\includegraphics[scale=0.35]{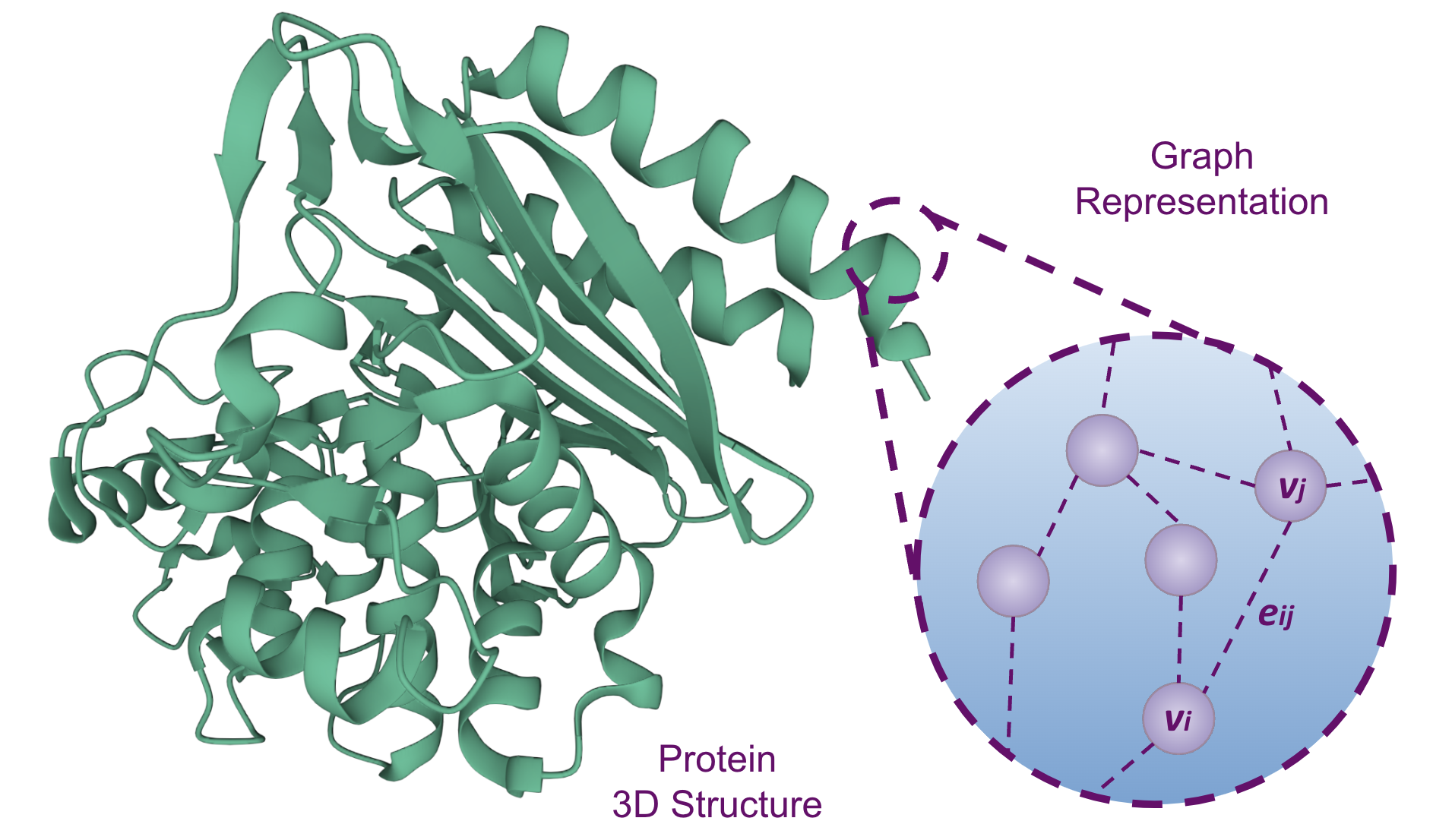}
\caption{Protein 3D structure visualization (protein \textit{\textbf{1GA0}} from the PDB shown here with its secondary structure components). The graph representation of the protein consists of nodes \(v\) (atoms) and edges \(e\) (covalent bonds between the atoms). The edge between two nodes \(v_{i}\) and \(v_{j}\) is defined as \(e_{ij}\).}
\label{fig:fig1}
\end{center}
\end{figure}

Figure \ref{fig:fig1} illustrates the conversion from a protein 3D structure to a graph representation, where each node represents an atom and each edge represents a bond. We adopt the graph representation used in \textit{Pytorch Geometric} \cite{pyg}. Each protein is represented as a graph object that contains the following matrices:

\begin{itemize}
\item The matrix of the nodes features, with dimension \textit{[number of nodes, number of node features]};
\item The matrix of the 3D coordinates of the nodes, with dimension \textit{[number of nodes, 3]};
\item The matrix of the edge features, with dimension \textit{[number of edges, number of edge features]};
\item The target node feature vector to predict, with dimension \textit{[number of nodes]}.
\end{itemize}

\subsection{Node Features}

Several node features of the protein atoms are included in the graph object. The categorical node features are represented as one-hot encoders, while the numerical features are normalized and scaled.

\paragraph{Atom type} The atom element (here referred as atom type) is included in the set of node features as a one-hot encoder. The majority of heavy atoms in the protein structure are carbon (C), oxygen (O), and nitrogen (N). The hydrogen atoms are very light compared to other atoms and are rarely modeled by crystallographers.

\paragraph{Atom relative location} Every atom from the PDB also contains information about its relative location (also called \textit{locant}) in the residue. Typically, atoms closer to the backbone will present less mobility than the ones further in the side chain. We include this information as a one-hot encoder.

\paragraph{Atom degree} The atom packing in the 3D space (i.e. density) can also influence their ability to move. Most non-covalent interactions (e.g. hydrogen bonds, charged interactions) are the strongest within range 2 and 5$\mathring{A}$. The atom degree is computed as the number of atom neighbors present in the 5$\mathring{A}$ radius of the 3D space and normalized. 
\paragraph{Residue type} Since properties like size, hydrophilicity, charge, and aromaticity can potentially influence side chain flexibility, a one-hot encoder is included for residue type, and each of the 20 natural amino acids is a possible value. 

\subsection{Edge Features}

Two edge feature categories were included in the graph object. The categorical features are incorporated as one-hot encoders while the numerical features are normalized and scaled.

\paragraph{Covalent bond type} For the covalent bond between two atoms (nodes), we incorporate the type of bond (i.e. single, double, triple, delocalized) as a one-hot encoder.
\paragraph{Distance:} Given two connected nodes $\mathbf{v_i}$ and $\mathbf{v_j}$ with 3D coordinates $\mathbf{p_i}$ and $\mathbf{p_j}$, we compute the Euclidean distance as $\mathbf{\sqrt{(p_i - p_j)^2}}$ and incorporate it into the edge features.

\begin{table}[t]
\centering
\caption{Quantitative comparison of the Apo/Holo dataset (training) and the Kinase dataset (testing) in terms of number of proteins, number of residues, number of atoms, crystallography resolution and average B-factor.}
\label{tab:table1}
    \small
    \begin{tabular}{cccccccc}
    \hline
    \textbf{Dataset} & \textbf{Number of } & \textbf{Average number} & \textbf{Average number} &  \textbf{Total number} & \textbf{Total number}  & \textbf{Average} & \textbf{Average}\\
    \textbf{} & \textbf{proteins} & \textbf{ of residues} & \textbf{of atoms} &  \textbf{of residues} & \textbf{of atoms} & \textbf{resolution} & \textbf{B-Factor}\\
    \hline
    \textbf{Apo/Holo} & 2661 & 1004 & 5452 & 2.6M & 14.5M & 1.89 $\mathring{A}$ & 31.3 $\mathring{A}^2$\\
    \textbf{Kinase} & 4313 & 663 & 3886 & 2.9M & 17.4M & 1.96 $\mathring{A}$ & 36.9 $\mathring{A}^2$\\ \hline
    \end{tabular}
\end{table}


\subsection{General Notation}

\paragraph{} For each protein of the dataset, we consider its graph representation $G = (V, E)$, where $V$ are the nodes representing its atoms and $E$ are the edges representing the covalent bonds between atoms. We define $X \in \mathbb{R}^{N \times C} $ as the node feature matrix, where $N$ is the number of nodes and $C$ is the number of features for each node. Similarly, we can also define $E \in \mathbb{R}^{M \times F} $ as the edge feature matrix, where $M$ is the number of edges and $F$ is the number of features for each edge.
The general framework of GNN layers consists of two main operations: (i) aggregate the information of neighboring nodes via a function $\phi_A$ and (ii) update the node embeddings by combining the incoming aggregated neighbor information to the current node embedding via a function $\phi_U$. $\phi_U$ and $\phi_A$ in this work are learned using different neural network architectures. The input node feature matrix $X^0$ is updated after each GNN layer to generate the learned hidden node representations $X^k$ after layer $k$. For instance, in a simplified case where the GNN layer only uses previous node attributes (i.e. no edge attributes involved), the node embeddings $x_v^{k+1}$ at layer $k+1$ for node $v$ that has a set of node neighbors $N(v)$ could be expressed by the following formulations:
\begin{equation}
	m_v^{k+1} = \phi_A^{k+1}(x_w^{k}), w \in N(v)
\end{equation}
\begin{equation}
	x_v^{k+1} = \phi_U^{k+1}(x_v^{k},m_v^{k+1})
\end{equation}
The last hidden node representations can then be used for downstream tasks such as node regression or node classification. For example, for predicting a node property such as the B-factor, a learnable regression head function $\phi_H$ (typically a simple Multilayer Perceptron) maps the last hidden encodings of node $v$, $x_v^{k+1}$, to a prediction $\hat{y}_v$:
\begin{equation}
	\hat{y}_v = \phi_H(x_v^{k+1})
\end{equation}

Depending on the GNN architectures, the functions $\phi_A$ and $\phi_U$ can be implemented differently and make use of node embeddings and/or edge embeddings.

\subsection{Models}

\begin{table}[t]
\centering
\small
\caption{Summary of the main design differences between the graph neural network architectures implemented in our framework. For each architecture implemented, we indicate whether the model layers: (i) update the node embeddings, (ii) use the edge features to update the node embeddings, (iii) update the edge embeddings, (iv) use attention mechanisms, (v) use positional embeddings and (vi) use pooling mechanisms.}
\label{tab:table_nets}
    \begin{tabular}{ccccccc}
    \hline
    \multirow{2}{*}{\textbf{Architecture}} & \textbf{Node} & \textbf{Uses edge features} & \textbf{Edge} & \textbf{Attention} &  \textbf{Positional} & \textbf{Pooling} \\
    \textbf{} & \textbf{updates} & \textbf{for node updates} & \textbf{updates} & \textbf{mechanisms} &  \textbf{encodings} & \textbf{mechanisms} \\
    \hline
    \textbf{GCN} & $\checkmark$ & $\times$& $\times$ & $\times$ & $\times$ & $\times$ \\
    \textbf{GAT} & $\checkmark$ & $\checkmark$ & $\times$ & $\checkmark$ & $\times$ & $\times$ \\
    \textbf{MPN} & $\checkmark$ & $\checkmark$ & $\times$ & $\times$ & $\times$ & $\times$ \\
    \textbf{UNet} & $\checkmark$ & $\times$ & $\times$ & $\times$ & $\times$ & $\checkmark$ \\
    \textbf{TConv} & $\checkmark$ & $\checkmark$ & $\times$ & $\checkmark$ & $\times$ & $\times$ \\
    \textbf{EGNN} & $\checkmark$ & $\checkmark$ & $\times$ & $\times$ & $\checkmark$ & $\times$ \\
    \textbf{Meta} & $\checkmark$ & $\checkmark$ & $\checkmark$ & $\checkmark$ & $\times$ & $\times$ \\
    \hline
    \end{tabular}
\end{table}

In computational chemistry, it is not established which architectures are more suited for predicting B-factors. Similar to the use of GNNs to predict force fields for small molecules in \cite{Park2021}, certain architectures excel at the atomic prediction task for proteins. Therefore, we have selected a diverse set of GNN architectures to explore their capabilities and to build a model that can incorporate valuable elements from the explored architectures.

\begin{figure}
\begin{center}
\includegraphics[scale=0.37]{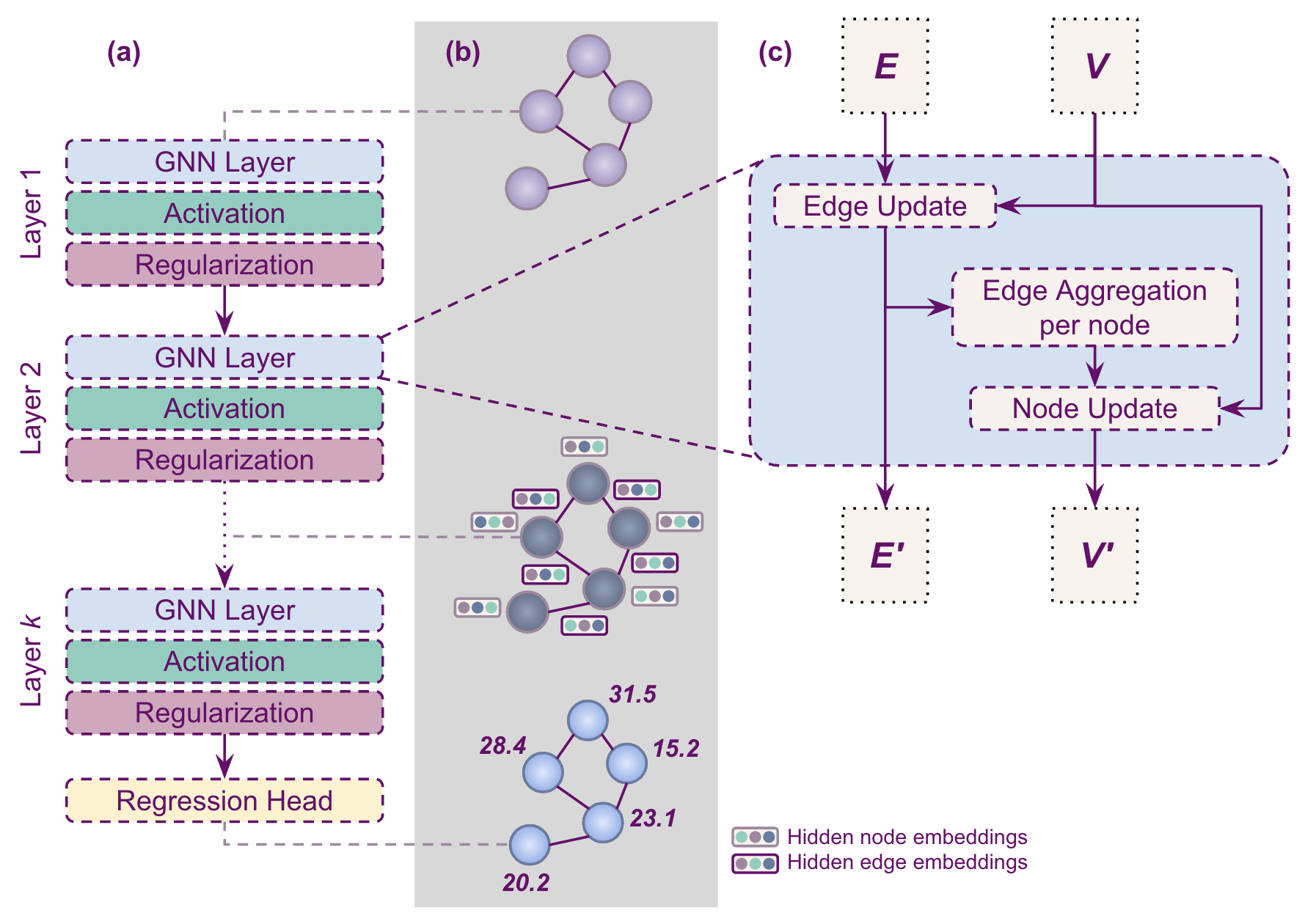}
\caption{Overview of a general graph neural network (GNN) architecture for the task of node prediction. \textbf{(a)} Each block contains a learnable GNN layer that updates nodes and/or edges representations, an activation function and a regularization (e.g. dropout, batch normalization). Skip connections are also typically implemented between consecutive blocks to help the training and improve the performance of the models. The regression head is typically a sequence of multi-layer perceptron (MLP) layers that map the last hidden node embeddings to the final node predictions. \textbf{(b)} A protein with input node and/or edge features (top) is typically fed into the network and goes through layers that learn hidden latent representations of its nodes and/or edges (middle) relevant to the training task. The output is a protein with node predictions such as the B-factor (bottom). \textbf{(c)} A general-purpose GNN layer (the one proposed by \cite{meta} is shown here as an example) is typically responsible for taking input matrices of node and/or edge embeddings \(V\) and \(E\), and generating updated embeddings \(V'\) and/or \(E'\). Depending on the choice of GNN layer, different mechanisms can be used to perform the aggregation and update operations.}
\label{fig:fig2}
\end{center}
\end{figure}

\paragraph{} Different GNN architectures from the literature were selected and adapted to the task of protein B-factor prediction. The diversity of the chosen architectures results in the exploration of different types of aggregation and update mechanisms. Each architecture design is based on different inductive biases of the graph data and can thus potentially learn and encode different embeddings. The seven models presented in this report fall into the following GNN architectures:
\begin{itemize}
\item \textbf{Convolutional} graph networks: the GCN architecture from \cite{gcn};
\item \textbf{Attentional} graph networks: the GAT architecture from \cite{gat} and the Transformer Convolutional layer from \cite{transformerconv};
\item \textbf{Message passing} graph networks: the MPN formulation from \cite{mpnn};
\item \textbf{Encoder-decoder} graph networks: the Graph-UNet architecture from \cite{graph-unet};
\item \textbf{Equivariant} graph networks: the EGNN architecture from \cite{egnn};
\item \textbf{General} multipurpose graph networks: the general graph network formulation from \cite{meta}.
\end{itemize}
Our implementations are based on \textit{PyTorch} and the \textit{PyTorch Geometric} package \cite{pyg}, unless otherwise stated. For all implementations, regularization (dropout and/or batch normalization) and residual connections between consecutive layers are implemented when suitable. A general overview of the models presented is illustrated in Figure \ref{fig:fig2}. A brief description and key differences of each architecture can be found in Table \ref{tab:table_nets}, and a more detailed description for all models except the Meta GNN model is given in Supplementary Equation \ref{appendix_math}, we describe the general notation of GNNs and the Meta model in the next section.

\subsection{Meta GNN - General Graph Networks}

\paragraph{} The general meta-layer proposed by \cite{meta} generalizes the concept of graph networks with three levels of features: node features, edge features and global graph features. Each graph network layer takes input node, edge and global features $V$, $E$ and $u$, and outputs a graph with updated features $V'$, $E'$ and $u'$. In the case where we only make use of nodes and edges features, three operations are needed to go from layer $k$ to layer $k+1$: compute updated edge attributes $e^{k+1}$ using edge attributes $e^{k}$ and its receiver and sender node attributes $v_{r}^{k}$ and $v_{s}^{k}$ (equation \ref{eq:meta1}), aggregate edge attributes per node to obtain $\hat{e}^{k+1}$ (equation \ref{eq:meta2}), compute updated node attributes $v^{k+1}$ using aggregated edge attributes $\hat{e}^{k+1}$ and node attributes $v^{k}$ (equation \ref{eq:meta3}). Figure \ref{fig:fig2} (c) illustrates the update mechanisms of the Meta layer. These steps can be formalized with the following equations for a given edge $i$ and a given node $j$, where $N^e$ is the number of edges, $N^n$ is the number of nodes and $E_j^{k+1}$ is the collection of edge vectors with receiver node $j$:

\begin{equation}
\label{eq:meta1}
	e_i^{k+1} = \phi_e^{k+1}(e_i^{k},v_{s_i}^{k},v_{r_i}^{k}), i \in \{1...N^e\}
\end{equation}
\begin{equation}
\label{eq:meta2}
	\hat{e}_j^{k+1} = \rho(E_j^{k+1}), j \in \{1...N^n\}
\end{equation}
\begin{equation}
\label{eq:meta3}
	v_j^{k+1} = \phi_v^{k+1}(\hat{e}_j^{k+1},v_j^{k}), j \in \{1...N^n\}
\end{equation}

In their simplest case, functions $\phi_e$ and $\phi_v$ can be implemented as learnable Multi-Layer Perceptrons (MLPs). In our implementation, we provide an edge update module $\phi_e$ that is a stack of MLP layers and 2 options of node update modules: a standard MLP $\phi_v$ and a $\phi_{v}$ that follows the node update mechanism of the GAT architecture (\cite{gat}). $\rho$  is an aggregation mechanism used to aggregate edge updates and is shared between updates. This architecture, unlike previous GNN layers presented (e.g. GCN, GAT, MPN), naturally enables the updating of the edge attributes after each layer. Moreover, it opens the door for the potential use of global features (e.g. protein family, protein size), and for a more fine-grained design of the aggregation and update mechanisms of each module. An overview of the main differences between the 7 architectures implemented is presented in Table \ref{tab:table_nets}.

\subsection{Experiments}
\paragraph{Dataset splits} The training was done on the Apo/Holo dataset for all models, using a random split of 85/15\% for the training and validation sets based on the same seed. The Kinase set is a test set to evaluate the performance of the best trained models. Apo/Holo are more general than the Kinase, so we expect to generalize to the Kinase when testing on them.

\paragraph{Evaluation Metrics} Based on the literature (\cite{jing2014research}, \cite{Bramer_2018}), we consider the following metrics for the evaluation of the performance on the node regression task: the \textit{Pearson correlation coefficient} (CC), the \textit{mean absolute error} (MAE), the \textit{mean absolute percentage error} (MAPE) and the \textit{root relative squared error} (RRSE). The CC measures the correlation between the target and predicted atomic B-factors, while the MAE, MAPE and RRSE measure the difference between the target and prediction values. Given a set of \(n\) nodes, a set of targets \(y\) and the corresponding set of predictions \(\hat{y}\), the performance metrics are defined as follows:

\begin{equation}
	CC = \frac {\sum_{i=1}^{n} (y_i - \bar{y})(\hat{y_i} - \bar{\hat{y}})}{\sqrt{\sum_{i=1}^{n} (y_i - \bar{y})^2\sum_{i=1}^{n}(\hat{y_i} - \bar{\hat{y}})^2}}
\end{equation}
\begin{equation}
\label{eq:mae}
	MAE = \frac {1}{n} \sum_{i=1}^{n} |y_i - \hat{y_i}|
\end{equation}
\begin{equation}
	MAPE = \frac {1}{n} \sum_{i=1}^{n} \frac{|y_i - \hat{y_i}|}{\max(\epsilon, |y_i|)}
\end{equation}
\begin{equation}
	RRSE = \sqrt{\frac{\sum_{i=1}^{n} (y_i - \hat{y_i})^2}{\sum_{i=1}^{n} (y_i - \bar{y})^2}}
\end{equation}
The best predictive models are expected to have a CC close to 1, and MAE, MAPE and RRSE values close to 0.

\paragraph{Training setup} Hyper-parameter search was performed on all 7 architectures for 10 epochs and the best models were selected based on the average CC metric over the validation data. Among the runs with train and validation CCs of greater than 0.6 and validation a MAPE lower than 0.5, we select the runs with the highest validation CC, since what is most important is to capture the trends of B-factor values within each protein. More details on the hyperparameter tuning process and a list of hyperparameters of each model can be found in Supplementary Section \ref{appendix_b}. All final models were trained for a maximum of 50 epochs, using the MAE loss (see equation \ref{eq:mae}) and the weighted Adam optimizer. A batch size of one was used (i.e. one protein) since some proteins are very large and can cause GPU memory issues. For each model, 3 different runs were executed, and the average metrics over these runs were reported. All models were trained on NVIDIA GPUs (models A100, V100 and/or RTX8000), using one GPU node per training session.

\paragraph{Training and Inference Time} Our models typically take around 2 minutes per epoch (using a batch size of one protein and a training set of almost $3k$ proteins) to train on a GPU. The inference speed is $\sim$37 proteins/s on a GPU and $\sim$15 proteins/s on a CPU. This allows us to potentially run predictions on more than $100k$ proteins per hour.

\section{Results}
\begin{figure}
\begin{center}
\includegraphics[scale=1]{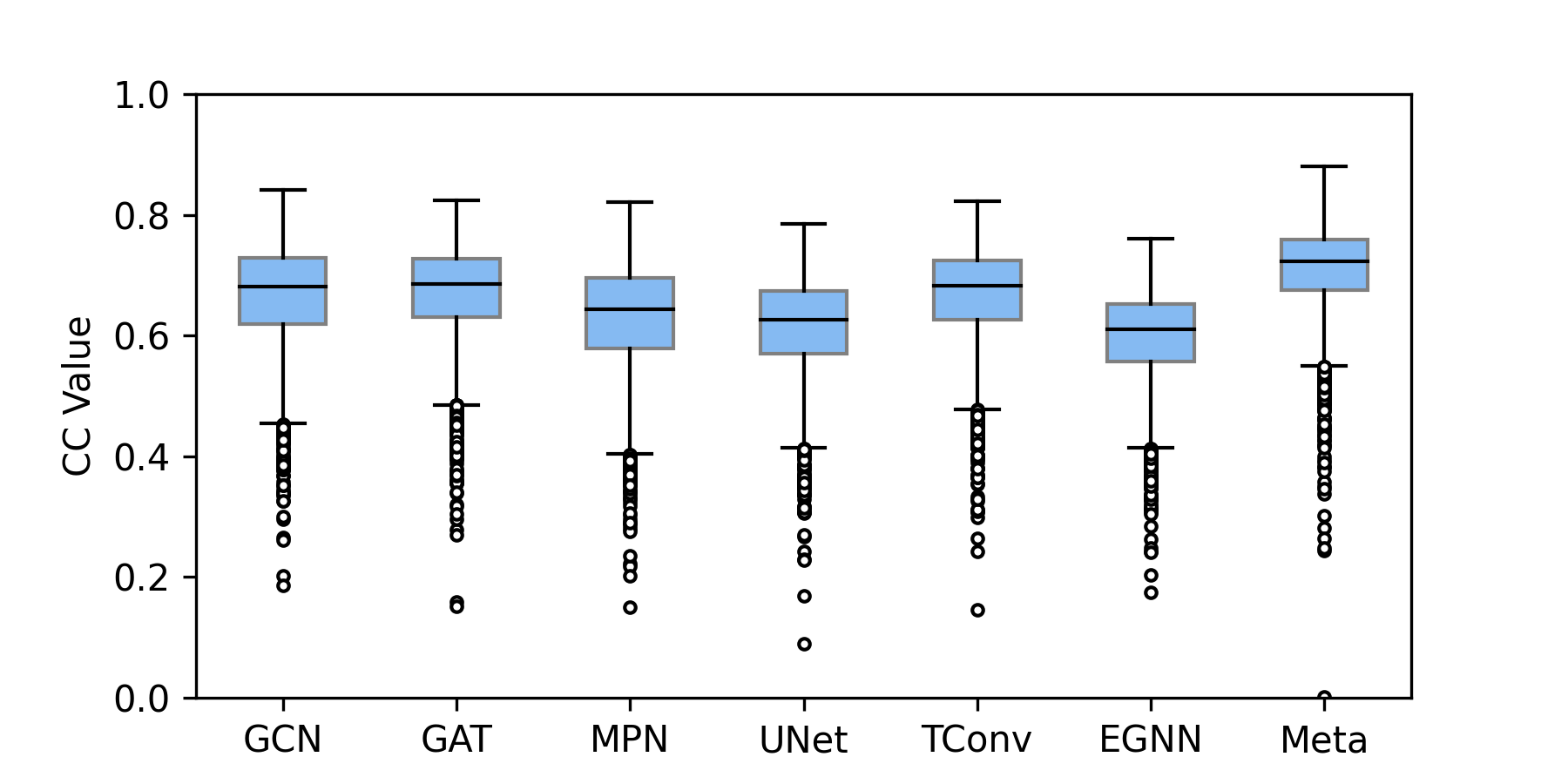}
\caption{Visualization of the Pearson correlation coefficient (CC) distributions of the proteins from the Kinase test set for each model. For each protein, the value reported is the average CC computed over the 3 runs done for each model. The blue box represents the interquartile range, the line in it is the median, and the dots are outliers.}
\label{fig:fig3}
\end{center}
\end{figure}

\paragraph{Performance metrics} The performance metrics of each model after hyper-parameter tuning on the training and validation sets are presented in Table \ref{tab:table2}. The models' performance metrics on the Kinase test dataset are presented in Table \ref{tab:table3}. As it can be observed in Table \ref{tab:table3}, the Meta-GNN implementation is the best-performing model on all metrics, obtaining a CC of $0.71$ as well as the lowest MAE, MAPE and RRSE values. Interestingly, the Meta-GNN is also the model with the second-lowest number of trainable parameters ($145k$). It is followed by the GCN, GAT and TConv implementations (CC of $0.67$). For a more detailed visualization of the protein CC distributions on the test set, see Figure \ref{fig:fig3}.

\begin{figure}
\begin{center}
\includegraphics[scale=0.24]{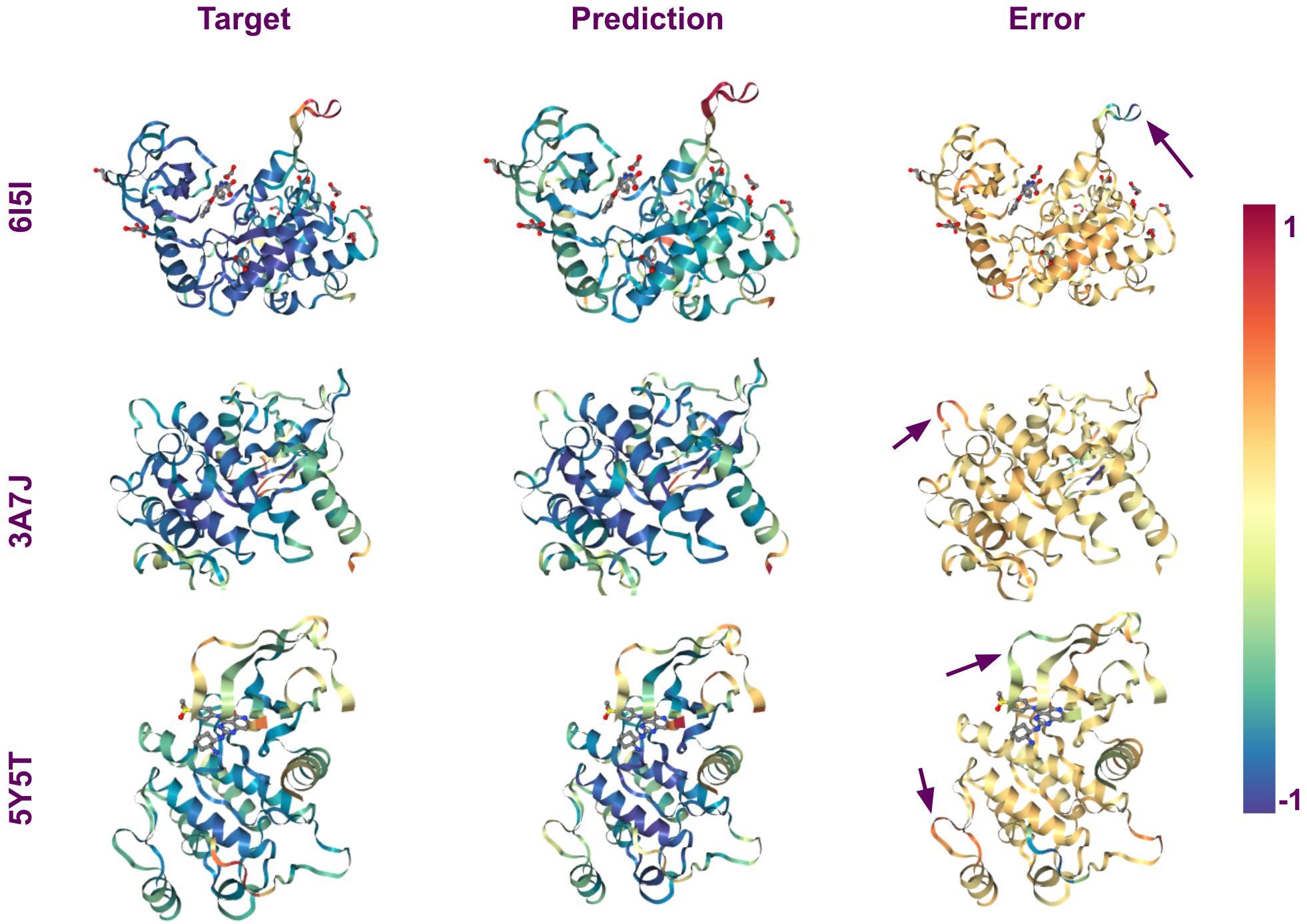}
\caption{Visualization of the B-factor predictions obtained on proteins \textit{\textbf{6I5I}} (CC of 0.88 \& MAE of 6.78), \textit{\textbf{3A7J}} (CC of 0.85 \& MAE of 4.74) and \textit{\textbf{5Y5T}} (CC of 0.84 \& MAE of 10.68) from the test set using the Meta model. For each protein, the target B-factor values (ground truth), the prediction B-factor values and the errors (computed as $prediction - target$) are projected into the 3D structure. All values are scaled between -1 and 1. Arrows highlight a few differences that can be observed in outer regions.}
\label{fig:fig4}
\end{center}
\end{figure}

To illustrate the predictions of the Meta model, we present three proteins from the test set and their projected predictions into the 3D structure in Figure \ref{fig:fig4}. In addition, a comparison between target and prediction B-factor distributions of a few selected proteins from the test set is presented in Supplementary Figure \ref{fig_histo}.

\begin{table}[h]
\centering
\caption{\textbf{Training} and \textbf{validation} performance metrics of  the highest performing configuration for each architecture (after hyper-parameter tuning) on the \textbf{Apo/Holo} dataset. The metrics reported are the Pearson correlation coefficient (CC), the mean absolute error (MAE), the mean absolute percentage error (MAPE) and the root relative squared error (RRSE). Experiments are repeated three times (using the same seed for the split) and values are presented in the form of \textbf{\textit{mean $\pm$ standard deviation}}. The best values of each column are highlighted in bold. For the CC, standard deviation values smaller than 0.01 are reported as 0.01 in the table for simplicity.}
\label{tab:table2}
{
\small %
\begin{tabular}
    {
    p{10mm}p{14.6mm}p{14.6mm}p{16.2mm}p{16.2mm}p{14.6mm}p{14.6mm}p{14.6mm}p{14.6mm}}
    
    \multirow{2}{*}{\textbf{Model}} & \multicolumn{2}{c}{\textbf{CC}} & \multicolumn{2}{c}{\textbf{MAE}} & \multicolumn{2}{c}{\textbf{MAPE}} & \multicolumn{2}{c}{\textbf{RRSE}} \\ \cline{2-9} 
    & \multicolumn{1}{c}{train.} & \multicolumn{1}{c}{val.} & \multicolumn{1}{c}{train.} & \multicolumn{1}{c}{val.} & \multicolumn{1}{c}{train.} & \multicolumn{1}{c}{val.} & \multicolumn{1}{c}{train.} & \multicolumn{1}{c}{val.} \\ \hline
    \textbf{GCN}
    & 0.67 $\pm$ 0.01
    & 0.68 $\pm$ 0.01 
    & 12.14 $\pm$ 0.02 
    & 12.66 $\pm$ 0.03
    & 0.48 $\pm$ 0.01
    & 0.55 $\pm$ 0.01
    & 1.21 $\pm$ 0.01
    & 1.26 $\pm$ 0.01 \\
    \textbf{GAT}
    & 0.67 $\pm$ 0.02
    & 0.68 $\pm$ 0.02
    & 12.13 $\pm$ 0.02 
    & 12.66 $\pm$ 0.06
    & 0.48 $\pm$ 0.02 
    & 0.55 $\pm$ 0.03
    & 1.20 $\pm$ 0.02 
    & 1.26 $\pm$ 0.02 \\
    \textbf{MPN}
    & 0.64 $\pm$ 0.02 
    & 0.65 $\pm$ 0.02 
    & 12.70 $\pm$ 0.68
    & 13.15 $\pm$ 0.52 
    & 0.47 $\pm$ 0.03
    & 0.53 $\pm$ 0.04
    & 1.24 $\pm$ 0.04
    & 1.28 $\pm$ 0.02 \\
    \textbf{UNet}
    & 0.62 $\pm$ 0.02
    & 0.62 $\pm$ 0.02 
    & 11.95 $\pm$ 0.17 
    & 12.45 $\pm$ 0.16 
    & \textbf{0.46} $\pm$ 0.02
    & 0.53 $\pm$ 0.02 
    & 1.20 $\pm$ 0.01 
    & 1.26 $\pm$ 0.02 \\
    \textbf{TConv}
    & 0.67 $\pm$ 0.01
    & 0.68 $\pm$ 0.01  
    & 12.05 $\pm$ 0.07 
    & 12.55 $\pm$ 0.01
    & \textbf{0.46} $\pm$ 0.02 
    & \textbf{0.52} $\pm$ 0.02
    & 1.20 $\pm$ 0.01 
    & \textbf{1.25} $\pm$ 0.01 \\
    \textbf{EGNN}
    & 0.61 $\pm$ 0.03 
    & 0.62 $\pm$ 0.03
    & 12.32 $\pm$ 0.21 
    & 12.89 $\pm$ 0.20
    & 0.50 $\pm$ 0.01 
    & 0.57 $\pm$ 0.01 
    & 1.26 $\pm$ 0.02 
    & 1.31 $\pm$ 0.02 \\
    \textbf{Meta}
    & \textbf{0.70} $\pm$ 0.01
    & \textbf{0.70} $\pm$ 0.01 
    & \textbf{11.48} $\pm$ 0.07 
    & \textbf{12.19} $\pm$ 0.07
    & 0.49 $\pm$ 0.01 
    & 0.57 $\pm$ 0.01
    & \textbf{1.19} $\pm$ 0.01 
    & 1.27 $\pm$ 0.01 \\ \hline
\end{tabular}%
}%
\end{table}

\begin{table}[h]
\centering
\small
\caption{Performance metrics on the \textbf{Kinase test} dataset. The metrics reported are the Pearson correlation coefficient (CC), the mean absolute error (MAE), the mean absolute percentage error (MAPE) and the root relative squared error (RRSE), as well as the number of trainable parameters. Experiments are repeated three times and values are presented in the form of \textbf{\textit{mean $\pm$ standard deviation}}. The best values of each column are highlighted in bold. For the CC, standard deviation values smaller than 0.01 are reported as 0.01 in the table for simplicity.}
\label{tab:table3}
\begin{tabular}{cccccc}

\textbf{Model} & \textbf{Number of param.} & \textbf{CC} & \textbf{MAE} & \textbf{MAPE} & \textbf{RRSE} \\ \hline
\textbf{GCN} & 95k
& 0.67 $\pm$ 0.01
& 13.65 $\pm$ 0.12
& \textbf{0.38} $\pm$ 0.01 
& 1.18 $\pm$ 0.01 \\
\textbf{GAT} & 853k
& 0.67 $\pm$ 0.02
& 13.62 $\pm$ 0.28 
& \textbf{0.38} $\pm$ 0.01
& 1.17 $\pm$ 0.01 \\
\textbf{MPN} & 412k
& 0.63 $\pm$ 0.04
& 14.83 $\pm$ 1.57 
& 0.39 $\pm$ 0.01 
& 1.27 $\pm$ 0.12 \\
\textbf{UNet} & 482k 
& 0.61 $\pm$ 0.03
& 13.50 $\pm$ 0.18
& \textbf{0.38} $\pm$ 0.01 
& 1.18 $\pm$ 0.01 \\
\textbf{TConv} & 553k
& 0.67 $\pm$ 0.01
& 13.70 $\pm$ 0.27 
& \textbf{0.38} $\pm$ 0.01 
& 1.18 $\pm$ 0.02 \\
\textbf{EGNN} & 248k
& 0.60 $\pm$ 0.04
& 13.67 $\pm$ 0.29 
& 0.40 $\pm$ 0.01
& 1.20 $\pm$ 0.03 \\
\textbf{Meta} & 145k
& \textbf{0.71} $\pm$ 0.01
& \textbf{12.30} $\pm$ 0.07 
& \textbf{0.38} $\pm$ 0.01 
& \textbf{1.09} $\pm$ 0.01 \\
\hline
\end{tabular}
\end{table}

\section{Discussion}

The GCN model is used as a baseline in this work since it is typically viewed as the simplest form of GNNs that can be applied to our task. The GCN model updates node features and considers a scalar constant like distance as an edge weight ( Supplementary Section \ref{gcn}).

Therefore, to include more information for the edges, models such as GAT, TConv, MPN, which aggregate edge features and use them to update node features, were also considered, since we have edge types too. We included the EGNN model, which includes encoded atom coordinates along with edge embeddings when updating node embeddings, because other models do not use this information.

We included UNet as it has a different aggregation mechanism compared to the other models, and operates based on pooling and unpooling node features (for more details refer to Supplementary Section \ref{unet}).
Finally, we combined features of the GAT architecture into the Meta model for updating both node embeddings and edge embeddings, making the Meta model the only model updating edge embeddings in this work.
The performance of the UNet model indicates that this architecture, which uses pooling layers, may not be well-suited for node regression tasks, as it was used for graph and node classification in the \cite{graph-unet}. This could be related to the lack of structural and edge information in the unpooling layers of the decoder (Supplementary Section \ref{unet}), and the loss of finer-grained details in the encoder, due to the pooling operation.

The EGNN model was selected because it is among the few models that embed node coordinates as features, and updating these coordinate embeddings at every layer is possible. Layers of the EGNN model can embed this information in a relative manner that is translation, rotation, and permutation equivariant, as proven in \cite{egnn}. The EGNN's performance on the test set is similar to UNet (Table \ref{tab:table3}). This model encodes more information compared to the others and considers symmetric features in the structures, which may not be helpful in the case of proteins, as structural symmetry does not necessarily result in similar B-factors. Note that the best configuration of hyperparameters for the EGNN model, based on the criteria specified in Supplementary Section \ref{appendix_b}, does not update nodes. It only updates the embeddings of the coordinates (Supplementary Table \ref{tab:table_tuning2}). It seems that the network learns to rely solely on the coordinates and propagates and updates the coordinate embeddings. This is similar to what UNet does, as distances to connected neighbors are added to node embeddings and propagated.

TConv, GAT, and MPN are all close in architecture and how they encode node and edge features. MPN has no attention mechanism when updating nodes by edge information, and this may justify its inferior performance compared to the other two, as it loses the original context, such as the distances between nodes, after every update.
While TConv and GAT perform similarly (Table \ref{tab:table_nets}), the performance of these architectures is not better than that of the GCN, and we can conclude that attention is not helpful for updating node features using edge features. This could be due to the fact that considering distances as edge weights in GCN is sufficient, and applying attention to it does not provide more context. However, we later see that having attention when updating edge embeddings is helpful for the Meta-GNN model.

So far, we have observed the importance of retaining distances in the node features. As GCN, GAT, and TConv all have different mechanisms for retaining this information, while UNet and MPN lose this information at every update. Now, we are curious to see how retaining and updating the distance information at the node and edge embeddings can affect the performance. This brings us to the Meta-GNN model.
In the Meta-GNN model, edges are updated after each layer, and this model is the only model in this work that updates edge features (Table \ref{tab:table_nets}) as well as node features. It seems that updating the edge features in this model leads to better performance compared to the other models, since this is the main difference between the Meta-GNN model and other models. The GAT model's node embedding layers are used in the Meta-GNN model so that the attention mechanism can be used to update node features, which gives better performance than using a simple linear aggregation.

The Meta-GNN model has a CC of 0.71 on the test set, which is higher than the best CC value (0.69) found in the literature (\cite{jing2014research}, \cite{Bramer_2018}), and it is tested on a much larger amount of data compared to the previous works, where they have used 474 proteins in \cite{jing2014research} and 364 proteins in \cite{Bramer_2018}. The prediction task that we considered here is per atom, not per residue, which is a much finer-grained task with higher degrees of freedom. We refrained from employing any feature extraction or descriptors generated by third-party software, relying solely on the information derived from the protein's structural characteristics. 

We believe that this work has many potential applications:
\begin{itemize}
    \item {
    The models provided can continually be re-trained and/or fine-tuned on new PDB structures deposited online to improve their performance and generalization.
    } 
    \item { Atomistic learning and prediction allows for ligand, co-factor, etc. B-factor prediction. }
    \item{
    Our models can serve as a fast way to obtain B-factor estimations of homology modeled protein structures such as those generated by AlphaFold.
    }
    \item{
    Our flexible and modular framework can be easily adapted to the prediction of other properties of interest (eg. thermostability, surface exposure, surface electrostatics and charge distribution) with minimal changes.
    }
\end{itemize}

\subsection{Limitations}
 When using the models for protein B-factor prediction in this work, one should consider the following limitations:
\begin{itemize}
    \item The models in the Protein Data Bank (PDB) can have problematic B-factor values, such as missing values, very high values (above 80), uniform values for all atoms, or significant variation between bonded atoms. These issues can be due to  over-fitting in crystallography models or may accurately represent disorder caused by a complex mix of protein dynamics and static disorder (e.g., from crystal imperfections). When lower-resolution structures have multiple conformations, B-factors may increase to account for uncertainty.

    To study the impact of protein dynamics on B-factors, it's crucial to avoid models with such artifacts as they can obscure useful information. Instead, training models on B-factors primarily reflecting thermal motion might be a viable approach. Then, comparing predicted and modeled B-factors in problematic structures can help identify the root cause of B-factor issues. Analyzing trends in B-factors from high-resolution data to medium-high resolution data may provide insights into the relationship between B-factor analysis and static and dynamic aspects of protein structure.
        
    \item Challenges can arise due to the experimental nature of the measurements: (i) many additional factors such as crystal defects, large-scale disorder, and diffraction data quality can deteriorate the overall data quality (\cite{carugo2022b}) and (ii) the values can vary significantly depending on the crystallography resolution (\cite{carugo2018large}) and between protein structures. A more in-depth curation of the PDB data/proteins used for training the models may be necessary to avoid learning from biased experimental data.
    \item Substantially different B-factors can be encountered for identical proteins at high resolution, but in different crystal packing and lattice contacts. Our models do not consider the possible interactions between the protein surface and the neighboring molecules/proteins in the crystal lattice. These interactions between proteins and neighboring elements can lead to lower B-factor values at the interface between the two, which the model cannot take into consideration, as it only has information about the protein of interest. 
    \item Among the models used here, only the EGNN model considers the coordinates of the atoms, which can be used to get an insight into the physical and chemical properties. None of the models can extract information on the Dihedral angles directly, these angles can potentially influence B-factors.
    \item We only consider edges between atoms that have a bond between them, not taking into account the non-covalent interactions between elements of the protein 3D structure. Non-covalent interactions between atoms can also influence their B-factors; however, adding them will significantly increase the computation overhead.
    \item The training and test data used in this work, although larger than previous works such as  \cite{jing2014research} and \cite{Bramer_2018}, are still a small representative of all known protein structures.
    \item We have mainly worked with single representatives of protein classes, while one could benefit from larger training data by allowing multiple representations of the same protein family.
\end{itemize}

\section{Conclusion and Future Work}

We presented here for the first time a deep learning framework predicting atomic protein B-factors based on their 3D structure using graph neural network architectures. Past work done on protein B-factor prediction has the following limitations: (i) they are trained and/or tested on very small datasets, (ii) they rarely reach CC values larger than 0.6 and (iii) they use traditional machine learning approaches that do not make use of the rich 3D graph structure of proteins. Our methodology addresses these limitations by providing a GNN model (Meta-GNN) that reaches an average correlation coefficient of $0.71$ on an unseen test set of more than $4k$ proteins ($17M$ atoms). We believe that this work can be a promising step towards atomic property prediction tasks on proteins, which can contribute to drug discovery research and applications such as protein misfolding.


This work also opens the door for many future ideas. We list a few of them here in an effort to motivate new discussions and increase the reach of the framework implemented:

\begin{itemize}
    \item \textbf{Processing Anisotropic B-factors:} Anisotropic B-factors \cite{yang2009comparisons} take the refinement process a step further by acknowledging that atoms can vibrate differently along different axes. Instead of assuming an equal level of vibration in all directions, anisotropic B-factors use a more sophisticated model that considers distinct vibrational behavior along the principal axes (x, y, and z) for each atom. It is interesting to see how the models developed and trained in this work will perform on predicting the anisotropic B-factors.
    \item \textbf{Novel GNN design based on the Meta-GNN formulation:} The promising performance of the modular Meta-GNN model leads us to believe that the next iteration of this work could consist of designing a novel architecture more adapted to the specific task of atomic property prediction on proteins. Interesting design ideas from various architectures could then be incorporated into a novel GNN design: positional encodings (EGNN), attention formulations (GAT, TConv), aggregation, and update mechanisms of the nodes, edges and global embeddings.
    \item \textbf{Hierarchical GNN approaches:} Hierarchical approaches that exchange information between multiple levels of protein structure (e.g. between atomic level and residue level) in order to capture these interactions would be interesting to explore (\cite{fey2020hierarchical}, \cite{somnath2021multi}). These multi-level approaches would be relevant for large macromolecules such as proteins that have primary, secondary, tertiary, and even quaternary structures. 
    \item \textbf{Larger input features space:} Due to the limited time-frame of this work, a broader exploration of node and/or edge features to incorporate into our data pipeline was not possible. More information on the crystal lattice can be added. Examples of potential features that could be added by using computational chemistry tools are angles, secondary structure information (e.g. alpha helices, beta sheets), and non-covalent interactions (e.g. hydrogen bonds).
    \item \textbf{Other losses and/or performance metrics:} It would be relevant to use other loss formulations that could capture the distribution differences between predictions and targets better. Metrics such as the Wasserstein distance could potentially help in that regard.
    \item \textbf{RNA and DNA representation:} It would be interesting to apply the GNN architectures in this work on tasks on RNAs and DNAs, to see how expressive they are on these types of data.
\end{itemize}

\section{Data Availability}
The proteins are retrieved from the PDB (\url{https://www.rcsb.org/}) . The code names are listed for the training set at \url{https://gitlab.com/congruencetx/pfp/-/blob/main/data/trainval_codes.txt} and for test set at \url{https://gitlab.com/congruencetx/pfp/-/blob/main/data/test_codes.txt}.

\section{Code Availability}
All the code developed in this project has been made freely available at \url{https://gitlab.com/congruencetx/pfp}.

\section{Competing interests}
The authors declare no competing interests.

\section{Supplementary Information} \label{appendix_math}

\subsubsection{GCN - Graph Convolutional Networks} \label{gcn}

\paragraph{} Graph Convolutional Networks are one of the simplest architectures for node classification tasks. For a GCN model with $K$ layers, the intermediate layers perform the following operations to update node embeddings:
\begin{equation}
	x_v^{k+1} = \phi_U^{k+1}(x_v^{k}, \sum_{w \in N(v)} c_{vw}x_w^{k}), k=1,\dots,K-1
\end{equation}
where $x_v^{k+1}$ is the node embedding for node $v$ at step $k+1$ that has a set of neighbors called $N(v)$. The function $\phi_U$ combines the neighboring features of node $v$ with its own features $x_v^{k}$ to compute the updated feature embeddings $x_v^{k+1}$. A set of constants $c_{vw}$ can be used to weigh every neighbor differently during the aggregation operation. 

A GCN layer does not make use of possible edge features, but it can be seen as a simple and efficient way to learn node representations based on its local neighbors. For instance, after 3 GCN layers, a given node $v$ will have a new hidden representation that captures all the previous hidden representations up to its 3-hop neighbors (i.e. nodes that it can reach with 3 edges or fewer) (see \cite{sanchez-lengeling2021a}). Our model implementation offers two options for regression heads: a standard MLP or a GCN layer that directly maps to the final node prediction value.

\subsubsection{GAT - Graph Attention Networks}

\paragraph{} In Graph Attention Networks (\cite{gat}),  the representation of a given node is updated based on the linear combination of its neighbors' representations. The shared weights $\theta$ in the linear combination are a set of learnable parameters known as the attention coefficients. 

In the GAT formulation, the attention mechanisms called $\alpha_{vw}$  for a given node $v$ with neighbors $w$ are computed as a $softmax$ over the concatenation of its features $x_v$, its neighbors' features $x_w$ and the edge features $e_{vw}$ between them transformed by the aggregation layer $\phi_A^k$, which is a linear layer with \textit{LeakyReLU} in our work.

A GAT layer $k+1$ will update the embeddings $x_v^{k+1}$ of node $v$ by: (i) aggregating the neighbors' features and their edge features via the attention mechanism and (ii) combining them with the previous embeddings $x_v^{k}$ of node $v$:
\begin{equation}
\label{eq:gat}
\alpha_{vw} = softmax(\phi_A^k( \theta^k x_v^{k},\theta^k x_w^{k},\theta^k e_{vw}))
\end{equation}
\begin{equation}
	x_v^{k+1} = \phi_U^{k+1}(\alpha_{vw} \theta^k x_v^{k},\sum_{w \in N(v)} \alpha_{vw} \theta^k x_w^{k})
\end{equation}

\subsubsection{TConv - Graph Transformer Networks}

\paragraph{}The Unified Message Passing (UniMP) model presented in \cite{transformerconv} proposes a graph transformer convolutional layer (which we will name TConv here for simplicity) that is another flavor of attention-based GNNs. The general formulation is very similar to the GAT layer, but the computation of the attention in the TConv layer is done similar to the \cite{vaswani2017attention}. A TConv layer $k+1$ will update the embeddings $x_v^{k+1}$ of node $v$ with the following equation:
\begin{equation}
K=\phi_k (x^k_v)
\end{equation}
\begin{equation}
Q= \phi_q( x^k_w +  e_{vw})
\end{equation}
\begin{equation}
\alpha_{vw} = softmax( QK^T/\sqrt{d})
\end{equation}
\begin{equation}
	x_v^{k+1} = \phi_U^{k+1}(K,\sum_{w \in N(v)}\alpha_{vw}Q)
\end{equation}
Where $\phi_k$ , and $\phi_q$ are networks for updating $K$ and $Q$ and d is the dimension of the embeddings.

\subsubsection{MPN - Message Passing Networks} \label{mpn}

\paragraph{} Message passing networks (\cite{mpnn}) provide a more general implementation of a GNN layer, standardizing the concept of message passing between a node and its neighbors. For every node $v$ of the graph, a message $m_v^{k+1}$ is computed by applying a differential function such as Multi Layer Perceptrons to  the concatenation of node features $x_k, x_w$ and edge features $e_{vw}$ . The aggregated message is then used to update the embeddings $x_v^{k+1}$.  The MPN operations can be expressed as follows:
\begin{equation} \label{eq:mpn_message}
	m_v^{k+1} = \sum_{w \in N(v)}\phi^k (x_v^{k},x_w^{k},e_{vw})
\end{equation}
\begin{equation} \label{eq:mpn_update}
	x_v^{k+1} = \phi_U^{k+1}(x_v^{k},m_v^{k+1})
\end{equation}

The $\phi$  in our case is a linear layer. There is a connection between  GAT, TConv and MPN, which is the way node and edge information are combined and transformed and the use of attention mechanism in GAT and TConv, and lack of it in MPN.

Unlike the original implementation (designed for small molecules), we do not use weight sharing between the different MPN layers, but rather learn different layers for each block to get more expressivity. Moreover, we drop the readout step since we are interested in a node level task rather than graph task.

\subsubsection{EGNN - Equivariant Graph Networks} \label{egnn}

\paragraph{} The Equivariant Graph Networks introduced in \cite{egnn} make use of coordinates of the nodes and can be used when coordinates are available. We adapt the implementation from \url{https://github.com/lucidrains/egnn-pytorch}. The embedding used for the coordinates in this model is translation, rotation, and permutation equivariant. In each layer, coordinates can be embedded using their relative distances as in Equation \ref{eq:egnn_position_update}.

\begin{equation}
\label{eq:egnn_position_update}
	m_{vw}^k = \phi_A (x_v^k , x_w^k, ||p_v^k - p_w^k||^2, e_{vw})
\end{equation}
\begin{equation}
	p^{k+1}_v = p^k_v + \frac{1}{M+1} \sum_{w \in N(v)} (p^k_v - p^k_w ) \phi_P (m_{vw}^k)
\end{equation}
\begin{equation}
	x^{k+1}_v = \phi_U (x_v^k , \sum_{j \neq i} m_{vw}^k )
\end{equation}
Here $\phi_P$ is the network that embeds coordinates of nodes with respect to its neighbors and the edge features, $p_i$ are the coordinates, and $e_{vw}$ are the edge attributes.
By adding positional encoding we make sense of the atoms' relative position in the protein space (instead of having to add edge distances between every atom of the graph).

\subsubsection{Graph U-Net - Encoder-Decoder Graph Networks} \label{unet}

\paragraph{} The graph U-Net architecture (\cite{graph-unet}) is directly inspired by the success of the original U-Net (\cite{unet}) on image segmentation tasks. It introduces implementations of learnable graph pooling and unpooling layers and integrates them into an encoder-decoder architecture for node classification tasks. The encoder path blocks of the network consist of a stack of GCN layers and pooling layers, where the node embeddings are updated and the hidden graph representation is down-sampled to only keep a portion of impactful nodes (\textit{top-k} pooling). In a symmetric fashion, the decoder path blocks take care of the up-sampling via unpooling layers to recover the original graph resolution, followed by a GCN layer. To combine the features of the encoder and decoder paths, skip connections are used between corresponding blocks to concatenate the node embeddings. For a U-Net architecture of $m$ total layers and a bottleneck (latent space) located at layer $k$, the encoder layers $X_E$ will range from $1$ to $k-1$, while the decoder layers $X_D$ will range from $k+1$ to $m$. Every block of the encoder path typically implements a GNN layer (GCN in the original paper\cite{graph-unet}) followed by a learnable pooling layer (where dropped nodes are selected via a learnable projection score). For instance, the encoder block of layer $k-1$ will first update the node embeddings $X_E^{k-2}$ to $X_E^{k-1}$ using $\phi_{GNN}^{k-1}$, and then apply the learnable pooling function $\phi_{pooling}^{k-1}$ to obtain the down-sampled $\tilde{X}_E^{k-1}$ node matrix.
\begin{equation}
	X_E^{k-1} = \phi_{GNN}^{k-1}(X_E^{k-2})
\end{equation}
\begin{equation}
	\tilde{X}_E^{k-1} = \phi_{pooling}^{k-1}(X_E^{k-1})
\end{equation}
Similarly, every block of the decoder path implements the opposite operation of unpooling followed by a GNN layer. For instance, the decoder block at layer $k+1$ will first recover the node embeddings from $\tilde{X}_D^{k+1}$ to $X_D^{k+1}$ using $\phi_{unpooling}^{k+1}$, and then compute new node updates via $\phi_{GNN}^{k+1}$ to obtain the $X_D^{k+1}$ node matrix. In the decoder path, the GNN layers are applied on the concatenation of node embeddings $X_D$ and their corresponding embeddings $X_E$ from the encoder path (e.g. concatenation between $X_E^{k-2}$ and $X_D^{k+2}$, between $X_E^{k-1}$ and $X_D^{k+1}$, and so on).
\begin{equation}
	X_D^{k+1} = \phi_{unpooling}^{k+1}(\tilde{X}_D^{k+1})
\end{equation}
\begin{equation}
	X_D^{k+1} = \phi_{GNN}^{k+1}(X_D^{k+1},X_E^{k-1})
\end{equation}

\newpage
\section{Target and Prediction Distributions Examples from the Kinase Set} \label{appendix_a}

\begin{figure}[h]
\begin{center}
\includegraphics[scale=0.47]{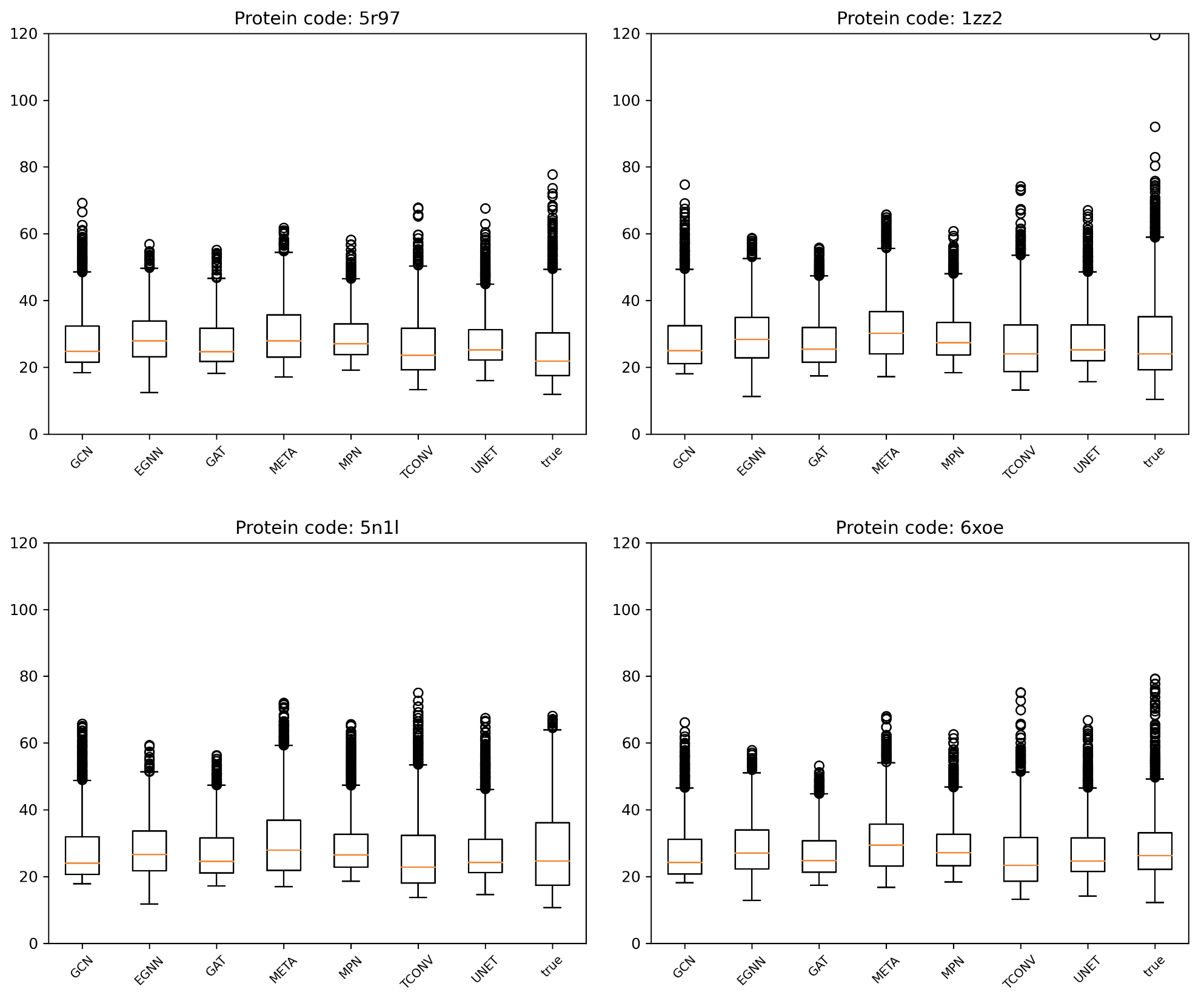}
\caption{Distributions of atomic B-factor targets (identified as \textit{true}) and predictions (identified by their model names) 
 from the trained models for a few proteins from the Kinase test dataset. For all plots, the y axis represents the B-factor values and the x axis represents the models/targets.}
\label{fig_histo}
\end{center}
\end{figure}

\newpage
\section{Hyper-parameter optimization details}\label{appendix_b}

A first filtering was done to keep experiments with train and validation CCs of greater than 0.6 and a validation MAPE lower than 0.5. Among the pool of candidates, the experiment yielding the highest validation CC was selected (further validated by observation of the learning curves). For all hyperparameter tuning experiments, the following parameters of the \textit{AdamW} optimizer were tuned: the learning rate (between $1 \times 10^{-5}$ and $1 \times 10^{-3}$), epsilon (between $0.1 \times 10^{-9}$ and $0.1 \times 10^{-7}$), and the weight decay (between $1 \times 10^{-5}$ and $1 \times 10^{-3}$). Table \ref{tab:table_tuning} presents the list of hyperparameters that were tuned for each of the implementations. A detailed list of all the parameters of the selected models and their values can be found in Table \ref{tab:table_tuning2} and at the \href{https://gitlab.com/r2414/revenir/external/mila-tempfactor/-/tree/main/experiments}{following link}.

\begin{table}[h]
\centering
\small
\caption{Hyperparameter tuning search space for each of the architectures trained. The number of trainable parameters and the values of the hyperparameters for the best-selected models are reported in Table \ref{tab:table_tuning2}.}
\label{tab:table_tuning}
\begin{tabular}{cl}

\textbf{Architecture} & \textbf{Hyperparameters tuned and search space}
\\ \hline
\textbf{GCN} 
& \begin{tabular}{@{}l@{}} 
Number of layers: integer in [1, 6] range.\\
Number of hidden channels: categorical among \{128, 256, 512\} \\ 
Use MLP head: \{True, False\} \\
Dropout: float in [0.1, 0.9] range.\\
Use residual connections: \{True, False\} \\
Use batch normalization: \{True, False\} \\
Use self-loops: \{True, False\} \end{tabular} \\ \hline
\textbf{GAT} 
& \begin{tabular}{@{}l@{}} 
Number of layers: integer in [1, 6] range.\\
Number of attention heads: integer in [1, 5] range.\\
Number of hidden channels: categorical among \{128, 256, 512\} \\ 
Activation function: categorical among \{ReLU, Tanh\} \\ 
Use MLP head: \{True, False\} \\
Use bias: \{True, False\} \\
Dropout: float in [0.1, 0.9] range.\\
Use residual connections: \{True, False\} \\
Use batch normalization: \{True, False\} \\
Use self-loops: \{True, False\} \end{tabular} \\ \hline
\textbf{MPN} 
& \begin{tabular}{@{}l@{}} 
Number of layers: integer in [1, 3] range.\\
Number of hidden channels: categorical among \{128, 256\} \\ 
Dropout: float in [0.1, 0.9] range.\\
Use residual connections: \{True, False\} \\
Use batch normalization: \{True, False\} \end{tabular} \\ \hline
\textbf{UNet} 
& \begin{tabular}{@{}l@{}} 
Depth of UNet architecture: integer in [1, 4] range.\\
Number of hidden channels: categorical among \{128, 256\} \end{tabular} \\ \hline
\textbf{TConv} 
& \begin{tabular}{@{}l@{}} 
Number of layers: integer in [1, 5] range.\\
Number of hidden channels: categorical among \{128, 256, 512\} \\ 
Number of attention heads: categorical among \{1, 2\} \\ 
Dropout: float in [0.1, 0.9] range.\\
Beta parameter: \{True, False\} \end{tabular}  \\ \hline
\textbf{EGNN} 
& \begin{tabular}{@{}l@{}} 
The number of layers: integer in [1, 6] range.\\
Dimension of the coordinate embedding network: categorical among \{16, 64, 128, 256, 512\} \\
A switch used to apply edge weights: categorical among \{0, 1\} \\
Update node features: categorical among \{True, False\} \\
Normalize node features: categorical among \{True, False\} \\
Normalize coordinates: categorical among \{True, False\} \\
Normalize coordinates scale: float in [0, 0.01] range.\\
Dropout: float in [0.1, 0.9] range.\\
Coordinates weights clamp value: float in [0.1, 1] range.\\
MLP activation: categorical among \{ReLU, Tanh\} \end{tabular} \\ \hline
\textbf{Meta} 
& \begin{tabular}{@{}l@{}} 
The number of blocks: integer in [1, 5] range.\\
Number of hidden channels of the head: categorical among \{128, 256\} \\ 
Dropout: float in [0.1, 0.9] range.\\
Use GAT layer instead of MLP for node updates: \{True, False\} \end{tabular} \\
\hline
\end{tabular}
\end{table}

\begin{table}
\centering
\small
\caption{Values of the hyperparameters of the best-selected models and resulting number of trainable parameters for each of the architectures.}
\label{tab:table_tuning2}
\begin{tabular}{clc}

\textbf{Architecture} & \textbf{Selected Hyperparameters} & \textbf{Number of params.}
\\ \hline
\textbf{GCN} 
& \begin{tabular}{@{}l@{}} 
Number of layers: 5.\\
Number of hidden channels: 128. \\ 
Use MLP head: True. \\
Dropout: 0.3569.\\
Use residual connections: True. \\
Use batch normalization: True. \\
Use self-loops: True. \\
\textit{Optimizer params:} \\
Learning Rate: 0.0008,
Epsilon: 1.3e-09,
$Beta_1$: 0.9, $Beta_2$: 0.999, 
Weight decay: 0.0002.
\end{tabular} 
& 95k \\ \hline
\textbf{GAT} 
& \begin{tabular}{@{}l@{}} 
Number of layers: 4.\\
Number of attention heads: 1.\\
Number of hidden channels: 512. \\ 
Activation function: ReLU. \\ 
Use MLP head: True. \\
Use bias: True. \\
Dropout: 0.1383.\\
Use residual connections: True. \\
Use batch normalization: True. \\
Use self-loops: True. \\
\textit{Optimizer params:} \\
Learning Rate: 0.0006,
Epsilon: 9.4e-09,
$Beta_1$: 0.9, $Beta_2$: 0.999,
Weight decay: 0.0007.
\end{tabular} 
& 853k \\ \hline
\textbf{MPN} 
& \begin{tabular}{@{}l@{}} 
Number of layers: 3.\\
Number of hidden channels: 128. \\ 
Dropout: 0.1994.\\
Use residual connections: True. \\
Use batch normalization: False. \\
\textit{Optimizer params:} \\
Learning Rate: 0.0008,
Epsilon: 2.7e-09,
$Beta_1$: 0.9, $Beta_2$: 0.999, 
Weight decay: 0.0009.
\end{tabular} 
& 412k \\ \hline
\textbf{UNet} 
& \begin{tabular}{@{}l@{}} 
Depth of UNet architecture: 4.\\
Number of hidden channels: 256. \\
\textit{Optimizer params:} \\
Learning Rate: 0.0007,
Epsilon: 4.7e-09,
$Beta_1$: 0.9, $Beta_2$: 0.999, 
Weight decay: 0.0004.
\end{tabular} 
& 482k \\ \hline
\textbf{TConv} 
& \begin{tabular}{@{}l@{}} 
Number of layers: 5.\\
Number of hidden channels: 128. \\ 
Number of attention heads: 2. \\ 
Dropout: 0.4257.\\
Beta parameter: False. \\
\textit{Optimizer params:} \\
Learning Rate: 0.0005,
Epsilon: 4.4e-09,
$Beta_1$: 0.9, $Beta_2$: 0.999, 
Weight decay: 0.0007.
\end{tabular}  
& 553k \\ \hline
\textbf{EGNN} 
& \begin{tabular}{@{}l@{}} 
Number of layers: 4.\\
Dimension of the coordinate embedding network: categorical among 16. \\
A switch used to apply edge weights: False. \\
Update node features: False.\\
Normalize node features: False. \\
Normalize coordinates: True. \\
Normalize coordinates scale: 0.0039. \\
Dropout: 0.4457.\\
Coordinates weights clamp value: 0.9349.\\
MLP activation: ReLU.\\
\textit{Optimizer params:} \\
Learning Rate: 0.0006,
Epsilon: 7.9e-09,
$Beta_1$: 0.9, $Beta_2$: 0.999,
Weight decay: 0.0001.
\end{tabular}   
& 248k \\ \hline
\textbf{Meta} 
& \begin{tabular}{@{}l@{}} 
Number of blocks: 5.\\
Number of hidden channels of the head: 128. \\ 
Dropout: 0.2987.\\
Use GAT layer instead of MLP for node updates: True. \\
\textit{Optimizer params:} \\
Learning Rate: 0.0007,
Epsilon: 2.5e-09,
$Beta_1$: 0.9, $Beta_2$: 0.999, 
Weight decay: 0.0001.
\end{tabular}  
& 145k \\
\hline
\end{tabular}
\end{table}

\newpage
\section{Explainability and Interpretability with GNN-Explainer}

There is growing interest in making large deep learning models more interpretable. This is even more relevant in fields such as healthcare or drug discovery, where users need to understand the final decisions made by such models. In an effort to bring more interpretability and explainability to graph neural networks, we present here a simple approach using GNN-Explainer \cite{gnn-explainer}. Training the GNN-Explainer on our GCN model as an example, which was trained on the protein B-factor prediction task, we attempt to answer the following question: which atom features are more impactful in the predictions made by the model? The GNN-Explainer is trained on a given node of a graph: it uses the GNN's computational graph to learn a graph mask that selects the important neighboring subgraph of that node. In parallel, it will also learn a feature mask that weights the node features in terms of impact on the final prediction for the node of interest.

\begin{figure}[h]
\begin{center}
\includegraphics[scale=0.45]{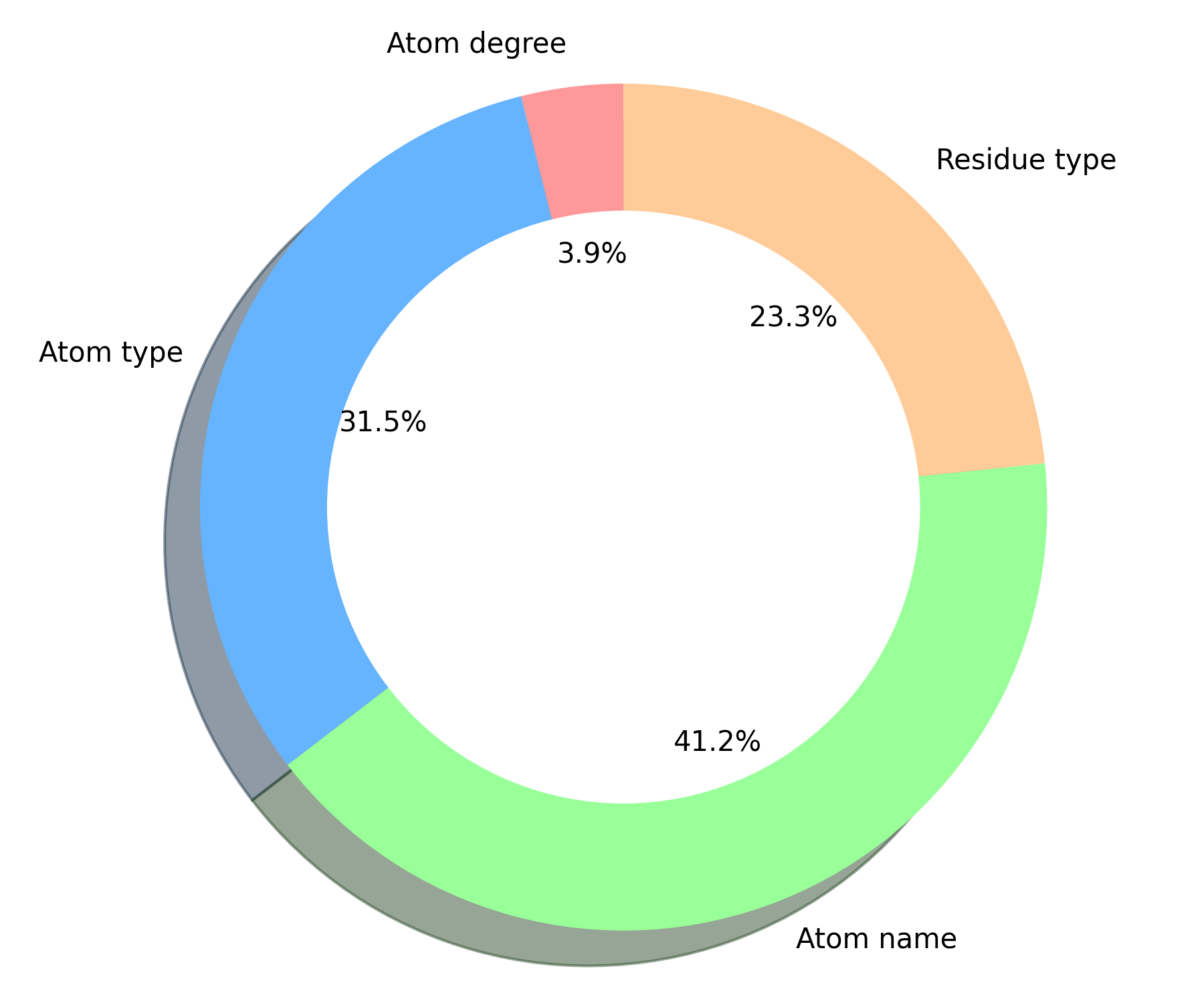}
\caption{Relative node feature importance (in \%) in protein \textbf{\textit{5Y9L}} in the B-factor prediction of selected atom \#400, computed by GNN-Explainer using our GCN model. Features are grouped by category: atom degree, atom type, atom name and residue type. Other atoms (nodes) output similar relative feature importance values per category.}
\label{fig:fig_residues}
\end{center}
\end{figure}

To demonstrate its potential, we train the GNN-Explainer over 200 epochs on a few selected nodes of protein \textbf{\textit{5Y9L}} while using our GCN model for B-factor predictions. In Figure \ref{fig:fig_residues}, we show a representative example of the impact of each node feature category on the B-factor prediction of a given node. In a more fine-grained analysis, we zoom in on the impact of each of the 20 residue type features on the prediction of the B-factor for a few atoms of protein \textbf{\textit{5Y9L}}. Results are presented in Table \ref{tab:table_residues}. Interestingly, different residue features have more impact (highlighted in bold) on the prediction depending on the atom. For instance, the prediction of atom \#100 is more strongly impacted by the presence (or absence) of the Tryptophan (W) feature channel.

\begin{table}[h]
    \centering
    \caption{Relative residue feature importance (in \%) on the B-factor prediction of a few atoms (nodes) of protein \textbf{\textit{5Y9L}}. Residues are presented in their one letter code. Values highlighted in bold indicate a higher impact on the prediction of that node.}
    \label{tab:table_residues}
    \begin{tabular}{P{12mm}P{3mm}P{3mm}P{3mm}P{3mm}P{3mm}P{3mm}P{3mm}P{3mm}P{3mm}P{3mm}P{3mm}P{3mm}P{3mm}P{3mm}P{3mm}P{3mm}P{3mm}P{3mm}P{3mm}P{3mm}}
    \hline
    \textbf{Atom ID} & \textbf{G} & \textbf{A} & \textbf{S}& \textbf{P}& \textbf{V}& \textbf{T}& \textbf{C}& \textbf{I}& \textbf{L}& \textbf{N}& \textbf{D}& \textbf{Q}& \textbf{K}& \textbf{E}& \textbf{M}& \textbf{H}& \textbf{F}& \textbf{R}& \textbf{Y}& \textbf{W}\\
    \hline
    \textbf{100} & 1.0 & 0.9 & 1.0 & 0.9 & 1.0 & 0.9 & 0.9 & 1.0 & 0.9 & 1.0 & 1.0 & 1.0 & 1.1 & 1.0 & 1.0 & 1.0 & 1.0 & 0.9 & 1.0 & \textbf{5.3} \\ 
    \textbf{200} & \textbf{2.4} & 0.9 & 0.8 & 0.9 & \textbf{2.7} & 0.8 & 0.9 & 0.8 & \textbf{4.2} & 0.9 & 0.9 & 0.9 & 0.9 & 0.8 & 1.0 & 0.9 & 0.9 & 0.9 & 0.9 & 0.8 \\ 
    \textbf{400} & 0.9 & 0.9 & \textbf{2.3} & 1.0 & 0.8 & \textbf{3.5} 
    & 0.9 & 0.9 & 1.0 & 0.8 & \textbf{2.2} & 0.9 
    & 1.0 & 0.9 & 1.0 & 0.9 & 0.9 & 0.9 
    & 0.8 & 0.8 \\ 
    \textbf{650} & 0.9 & 0.8 & 0.9 & 0.8 & 0.9 & 0.8 
    & 0.8 & 1.0 & \textbf{3.7} & 0.8 & 0.9 & 0.9 
    & 0.8 & 0.8 & \textbf{2.2} & 0.9 & 1.0 & 0.9 
    & 0.8 & 0.9 \\
    \textbf{900} & \textbf{1.9} & 1.1 & \textbf{2.6} & 1.0 & 1.0 & 1.0 
    & 0.9 & 1.0 & 1.0 & 0.8 & 0.9 & 0.9 
    & 1.1 & 0.9 & 0.8 & 1.0 & 1.0 & 1.0
    & 1.0 & 1.0 \\
    \textbf{1200} & 0.9 & 0.9 & 0.9 & 0.8 & 0.9 & 0.8 
    & 0.9 & 0.9 & 0.9 & \textbf{2.7} & 0.9 & 0.9 
    & 0.9 & 1.0 & 0.9 & 0.9 & 0.9 & 0.8
    & 0.8 & 0.9 \\
    \textbf{1500} & 0.9 & 0.9 & 1.0 & \textbf{4.3} & 0.9 & 0.9 
    & 0.9 & 1.0 & 0.9 & 0.9 & 0.8 & 1.0 
    & 0.9 & 1.0 & 0.9 & 1.0 & 1.0 & 1.0
    & 1.0 & 1.0 \\

    \hline
    \end{tabular}
\end{table}

\bibliographystyle{unsrtnat}
\bibliography{main}

\end{document}